\documentclass{article}

\PassOptionsToPackage{round}{natbib}
\usepackage[final]{neurips_2024}

\usepackage{amsmath}

\usepackage[utf8]{inputenc} % allow utf-8 input
\usepackage[T1]{fontenc}    % use 8-bit T1 fonts
\usepackage{hyperref}       % hyperlinks
\usepackage{cleveref}       % hyperlinks
\usepackage{url}            % simple URL typesetting
\usepackage{booktabs}       % professional-quality tables
\usepackage{amsfonts}       % blackboard math symbols
\usepackage{nicefrac}       % compact symbols for 1/2, etc.
\usepackage{microtype}      % microtypography
\usepackage{xcolor}         % colors
\usepackage{graphicx} 
\usepackage{tabularx}     % <— add this

\usepackage{ragged2e} % For \RaggedRight
\usepackage{algorithm}
\usepackage{algpseudocode}
\usepackage{algorithmicx}
\usepackage{caption}
\usepackage{longtable}      % Added for tables that may span pages
\usepackage{enumitem}       % Added for compact lists

\usepackage{tabularx,array}        % flexible tables
\newcolumntype{L}[1]{>{\raggedright\arraybackslash}p{#1}}
\newcolumntype{Y}{>{\raggedright\arraybackslash}X} % auto-stretch, ragged-right

\usepackage{booktabs,tabularx,threeparttable}
\newcolumntype{R}[1]{>{\raggedleft\arraybackslash}p{#1}}

\title{OpenMed NER: Open‑Source, Domain‑Adapted State‑of‑the‑Art
Transformers for Biomedical NER Across 12 Public
Datasets}

\author{
  Maziyar Panahi \\
  AI Engineer \\
  CNRS \\
  Paris, France \\
  \texttt{maziyar.panahi@cnrs.fr} \\
}

\begin{document}

\maketitle

\begin{abstract}
Named-entity recognition (NER) is fundamental to extracting structured
information from the >80\% of healthcare data that resides in
unstructured clinical notes and biomedical literature. Despite recent
advances with large language models, achieving state-of-the-art
performance across diverse entity types while maintaining computational
efficiency remains a significant challenge. We introduce \textbf{OpenMed NER}, a
suite of open-source, domain-adapted transformer models that combine
lightweight domain-adaptive pre-training (DAPT) with parameter-efficient
Low-Rank Adaptation (LoRA). Our approach performs cost-effective DAPT on a
350k-passage corpus compiled from ethically sourced, publicly available research
repositories and de-identified clinical notes (PubMed, arXiv, and MIMIC-III) using
DeBERTa-v3, PubMedBERT, and BioELECTRA backbones. This is followed by
task-specific fine-tuning with LoRA, which updates less than 1.5\% of
model parameters. We evaluate our models on 12 established biomedical NER
benchmarks spanning chemicals, diseases, genes, and species. \textbf{OpenMed NER achieves new state-of-the-art micro-F1 scores on 10 of these 12 datasets}, with substantial gains across diverse entity types. Our models advance the state-of-the-art on foundational disease and chemical benchmarks (e.g., BC5CDR-Disease, +2.70 pp), while delivering even larger improvements of over \textbf{5.3 and 9.7 percentage points} on more specialized gene and clinical cell line corpora (BC2GM and CLL, respectively).

This work demonstrates that strategically adapted open-source models
can surpass closed-source solutions. This performance is achieved with remarkable efficiency: training completes in under 12 hours on a single GPU with a low carbon footprint (\(<\!1.2\) kg CO\textsubscript{2}e), producing permissively licensed, open-source checkpoints designed to help practitioners facilitate compliance with emerging data protection and AI regulations, such as the \textbf{EU AI Act}.
\end{abstract}

\newpage

\tableofcontents

\newpage

\section{Introduction}

Recent advances in domain-specific language models, such as BioBERT~\citep{lee2019biobert}, ClinicalBERT~\citep{alsentzer2019clinicalbert}, and PubMedBERT~\citep{gu2021biomedbert}, have driven substantial progress in biomedical named-entity recognition (BioNER)~\citep{raza2022largescale,keraghel2024survey}. Yet, deploying these powerful models in real-world research and clinical settings requires overcoming persistent challenges like distribution shift across diverse corpora, tight compute budgets, and \textbf{the need for locally-deployable models to comply with stringent data protection regulations (e.g., GDPR, HIPAA)}. This paper demonstrates that a strategic combination of (i) \textbf{domain-adaptive pre-training} (DAPT)~\citep{gururangan2020dont}, (ii) \textbf{parameter-efficient fine-tuning} via Low-Rank Adaptation (LoRA)~\citep{hu2021lora}, and (iii) strong transformer backbones like \textbf{DeBERTa-v3}~\citep{he2021deberta} can yield state-of-the-art results while remaining computationally efficient and fully accessible.

The vast majority of information in biomedical research and healthcare is locked in unstructured text. Electronic Health Records (EHRs) alone are estimated to be \(\sim\!80\%\) unstructured: a category comprising clinical notes, research publications, drug labels, and patient narratives~\citep{li2021ehrnlp,yang2022gatortron}. Biomedical text also exhibits a high density of unique entities compared to general-domain corpora, making BioNER a particularly knowledge-intensive task~\citep{lee2019biobert,yadav2018survey}. Named-entity recognition therefore serves as a foundational technology for downstream applications in precision medicine, drug discovery, and clinical decision support~\citep{keraghel2024survey,raza2022largescale}.

This work, OpenMed NER, makes three primary contributions:
\begin{enumerate}
    \item We develop and release a suite of BioNER models that establish new state-of-the-art performance on 10 of 12 public benchmarks, outperforming proprietary systems with a lightweight and fully open-source framework.
    \item We provide a crucial ablation study quantifying the synergy between DAPT and LoRA, which together yield a \(2\text{–}4\%\) absolute F\(_1\) gain over using either technique in isolation.
    \item We release all final model checkpoints under the permissive Apache 2.0 license. This enables local, on-premise deployment, which is critical for applications handling sensitive health data and for facilitating compliance with data protection regulations like the EU AI Act.
\end{enumerate}

The biomedical domain poses linguistic challenges that extend far beyond
general-domain NER. Biomedical prose features rapid neologism,
systematic yet mutable nomenclatures, dense abbreviations, and
morphological variants that proliferate with each new discovery
\citep{lee2019biobert,lin2004maxent}.
Clinical entities further complicate matters through unclear boundaries,
frequent nesting, and cross-category aliases. For instance, the phrase
“upper respiratory tract infection (URTI)” can denote either a
\textsc{Disease} or a \textsc{Symptom} depending on context
\citep{alex2007nested}.
While early neural architectures such as BiLSTM-CNNs
\citep{chiu2016named} reduced feature engineering, they still struggled
with out-of-vocabulary terms and complex semantics.

The transformer era, initiated by BERT~\citep{devlin2018bert}, spurred
a new generation of domain-specific models. Variants like
BioBERT~\citep{lee2019biobert},
ClinicalBERT~\citep{alsentzer2019clinicalbert},
PubMedBERT~\citep{gu2021biomedbert}, and
BioELECTRA~\citep{kanakarajan2021bioelectra} proved that
\emph{domain-adaptive pre-training} (DAPT)~\citep{gururangan2020dont}
is critical for bridging the gap between general language understanding and
specialised biomedical semantics. More recent advances, including powerful
backbones like DeBERTa-v3~\citep{he2021deberta} and parameter-efficient
fine-tuning methods like LoRA~\citep{hu2021lora}, offer further
opportunities to push the state of the art.

Despite this rapid progress, critical gaps remain in BioNER research.
\textbf{First}, few head-to-head studies compare fully
\emph{open-source} models with proprietary cloud NLP APIs,
leaving the real-world performance and cost trade-offs opaque
and not always transparent in peer-reviewed literature.
\textbf{Second}, the synergy between modern backbones
(e.g., DeBERTa-v3) and parameter-efficient training has not been
systematically evaluated across the full spectrum of biomedical entity types.
\textbf{Third}, most prior work benchmarks on a narrow subset of
public datasets, often omitting clinically relevant corpora that test for
generalization, such as the BC4CHEMD~\citep{krallinger2015chemdner} and
Linnaeus~\citep{gerner2010species} datasets.

We address these gaps directly. By unifying efficient domain adaptation with
exhaustive, cross-dataset benchmarking, \emph{OpenMed NER} shows that
open-source solutions can equal or even exceed proprietary biomedical NER
systems.

\section{Related Work}\label{sec:related-work}

Progress in biomedical named-entity recognition (BioNER) mirrors three macro-trends in natural language processing: (i) the shift from feature-based methods to end-to-end neural encoders~\citep{chiu2016named,yadav2018survey}, (ii) the replacement of generic word embeddings with \textbf{domain-specific pre-trained language models (PLMs)}~\citep{lee2019biobert,gu2021biomedbert}, and (iii) the rise of \textbf{parameter-efficient adaptation, notably LoRA,} which updates less than 1.5\% of model weights while preserving performance~\citep{hu2021lora}. OpenMed NER is built upon these three pillars.

\subsection{Early Neural and Production-Scale Pipelines}

The BiLSTM-CNN-Char architecture of \citet{chiu2016named} represented a significant step forward for neural BioNER. Its re-implementation in the commercial \textbf{Spark NLP} library was significant for bringing these neural architectures to the scalable Apache Spark ecosystem, enabling large-scale, distributed inference across clusters~\citep{kocaman2022accurate}. Using more modern BERT-based models, Spark NLP has also reported strong, state-of-the-art results on several key benchmarks, demonstrating high performance on chemical and anatomical entity recognition tasks.

Other significant systems include \textbf{BERN2}, which couples multi-task learning with real-time entity normalization and reports an F\(_1\)-score of \textbf{92.70\%} on the Linnaeus species dataset~\citep{kim2021bern2}. Open-source toolkits such as \textbf{Stanza}~\citep{qi2020stanza} and \textbf{SciSpaCy}~\citep{neumann2019scispacy} also lowered the barrier to entry for researchers. However, while these toolkits were foundational, transformer-based PLMs have since been shown to outperform earlier architectures, often by significant margins on standard BioNER benchmarks~\citep{lee2019biobert,keraghel2024survey}. Furthermore, these libraries do not natively support modern, parameter-efficient fine-tuning techniques like LoRA, limiting their adaptability to new corpora and tasks.

\subsection{Domain-Specific Pre-trained Transformers}

Early biomedical PLMs demonstrated that \emph{continued} masked-language modeling on corpora like PubMed/PMC substantially boosts BioNER performance. \textbf{BioBERT} augmented BERT's training with 4.5B PubMed tokens and 13.5B PMC tokens, lifting performance on the NCBI-Disease benchmark from 86.7\% to \textbf{89.71\% F\(_1\)}~\citep{lee2019biobert}. \textbf{ClinicalBERT} replicated this strategy on MIMIC-III notes to better capture the distinct language of EHR narratives~\citep{alsentzer2019clinicalbert}.

A second wave of models showed the value of pre-training \emph{from scratch} on domain-specific data. \textbf{PubMedBERT} (also known as BiomedBERT) used only biomedical text and achieved new state-of-the-art results on the BLURB benchmark~\citep{gu2021biomedbert}. \textbf{BioELECTRA} adapted the more efficient replaced-token detection (RTD) objective to biomedical corpora, surpassing masked-language modeling variants on several tasks~\citep{kanakarajan2021bioelectra}. Concurrently, models like \textbf{BioMegatron} demonstrated the benefits of scale, using 345M parameters to reach an \textbf{88.50\% F\(_1\)} on BC5CDR-Disease~\citep{shin2020biomegatron}.

Knowledge-enhanced transformers add a third dimension by integrating structured knowledge. \textbf{KeBioLM} injects concepts from the Unified Medical Language System (UMLS) during pre-training, posting an \textbf{82.00\% F\(_1\)} on the challenging JNLPBA corpus~\citep{yuan2022kebiolm}. \textbf{SapBERT} uses a contrastive loss to align entity synonyms from UMLS, yielding strong performance on entity-centric tasks~\citep{liu2021sapbert}.

Collectively, the following trends define the landscape that OpenMed NER builds upon: continued domain-adaptive pre-training, from-scratch pre-training, alternative objectives, and structured knowledge alignment.

\subsection{Domain-Adaptive and Task-Adaptive Pre-training}

\citet{gururangan2020dont} introduced \emph{domain-adaptive pre-training} (DAPT) and \emph{task-adaptive pre-training} (TAPT), showing that a short, continued pre-training stage on unlabeled, in-domain data delivers significant gains. Subsequent biomedical studies confirm that DAPT yields average performance improvements of 2–4 percentage points on benchmarks like BLURB~\citep{gu2021biomedbert,kanakarajan2021bioelectra}. However, most prior work evaluates on eight or fewer datasets; our 12-corpus study is designed to close this evidence gap by providing a more comprehensive assessment of generalization.

\subsection{Parameter-Efficient Adaptation}

Full fine-tuning of a large PLM requires duplicating all 300M+ parameters for each new task. LoRA (Low-Rank Adaptation) offers a solution by inserting small, rank-decomposition matrices into the model and updating only these matrices, which often constitute less than 1.5\% of the total weights, while matching the performance of full fine-tuning~\citep{hu2021lora}. This approach has been successfully applied in the medical domain, demonstrating comparable performance to dense tuning while drastically reducing memory requirements~\citep{wu2023clinical_llama}. Frameworks like \textbf{AdapterHub} provide a unified interface for using such adapters with models from the Hugging Face ecosystem~\citep{pfeiffer2020adapterhub}, but systematic BioNER evaluations using these methods remain limited to a handful of datasets.

\subsection{Robustness and Specialized Systems}

Model robustness has become a central theme in recent BioNER research. For example, \textbf{ConNER} applies consistency training to improve performance on noisy data~\citep{li2023conner}. Stress tests reveal that even state-of-the-art models can fail to generalize, missing over 50\% of novel mentions of entities like COVID-19~\citep{vashishth2021generalizable}. A recent survey covering work from 2011–2022 shows that transformer-based methods now account for the majority of modern BioNER systems, far surpassing classical machine learning and CRF pipelines~\citep{keraghel2024survey}. Nonetheless, highly specialized tools remain competitive on their target domains, such as \textbf{LINNAEUS} for species recognition~\citep{gerner2010species}.

\subsection{Large Generative Biomedical LLMs}

Generative large language models (LLMs) are emerging as a complement to encoder-only PLMs. \textbf{BioGPT} achieved state-of-the-art results on PubMedQA and several relation extraction tasks~\citep{luo2022biogpt}, while the 8.9B-parameter \textbf{GatorTron} improved performance on multiple clinical NLP benchmarks~\citep{yang2022gatortron}. Instruction-tuned models like \textbf{BioInstruct} further enhance question answering and information extraction capabilities without task-specific heads~\citep{su2023bioinstruct}. However, these powerful autoregressive models remain computationally intensive for token-level tagging tasks like NER and are rarely benchmarked on classic BioNER corpora. For these reasons, OpenMed NER retains an efficient encoder architecture, which is the standard for this task, while remaining agnostic to future integration with generative models.

\section{Methods}
\label{sec:methods}

\subsection{Methodological Overview}
The OpenMed NER framework is built on a three-stage process designed to maximize performance while maintaining computational efficiency. The stages are:
\begin{enumerate}
    \item Domain-Adaptive Pre-training (DAPT): We first adapt general-purpose transformer backbones to the biomedical domain. Crucially, we perform this step using parameter-efficient LoRA adapters, which significantly reduces the computational cost compared to traditional full-model DAPT.
    \item \textbf{Task-Specific Fine-tuning:} The domain-adapted models are then fine-tuned on each of the 12 target BioNER corpora. During this stage, the base model parameters remain frozen, and only the LoRA adapters and a new token-classification head are trained.
    \item \textbf{Bayesian Hyper-parameter Optimization (HPO):} Finally, we use a systematic HPO search to identify the optimal configuration for fine-tuning, ensuring robust and high-performing models across all datasets.
\end{enumerate}
This approach follows the principles of "Don't Stop Pre-training"~\citep{gururangan2020dont} but integrates modern, efficient techniques~\citep{hu2021lora} to make the process accessible. We apply this method to three strong backbones: DeBERTa-v3~\citep{he2021deberta}, PubMedBERT~\citep{gu2021biomedbert}, and BioELECTRA~\citep{kanakarajan2021bioelectra}.

\subsection{Domain-Adaptive Pre-training (DAPT) with LoRA}
\label{sec:dapt}
The goal of DAPT is to infuse the models with specialized knowledge from the biomedical domain before task-specific fine-tuning. As detailed in Table~\ref{tab:dapt}, we continue pre-training on a diverse 350k-passage corpus using a masked-language modeling (MLM) objective. The entire DAPT process for a model completes in approximately 4 hours on a single NVIDIA A100-80GB GPU.

\begin{table}[ht]
\small
\centering
\caption{DAPT hyper-parameters for a single model on a single GPU.}
\label{tab:dapt}
\begin{tabularx}{\linewidth}{@{}L{0.22\linewidth}Y L{0.28\linewidth}@{}}
\toprule
\textbf{Component} & \textbf{Implementation Detail} & \textbf{Rationale} \\
\midrule
\textbf{Corpus} &
4-way mix, $\sim$350k passages (90M tokens): \newline
– 100k PubMed abstracts \newline
– 100k arXiv biomedical abstracts \newline
– 100k MIMIC-III sentences \newline
– 50k curated clinical trial descriptions from \href{https://clinicaltrials.gov/}{ClinicalTrials.gov} (filtered for 50–256 tokens) &
Balances formal scholarly language with clinical "bedside" narratives. The use of the public ClinicalTrials.gov repository ensures transparency and provides a rich source of structured biomedical text. \\
\addlinespace[0.6ex]
\textbf{Objective} &
Masked-Language Modeling (MLM) with 15\% dynamic masking via Hugging Face's \texttt{DataCollatorForLanguageModeling}. &
Standard DAPT objective, consistent with foundational models like BioBERT~\citep{lee2019biobert}. \\
\addlinespace[0.6ex]
\textbf{Adapter} &
LoRA with rank=16, $\alpha$=32, dropout=0.05, applied to query and value matrices in all attention layers. &
Updates $<$1.5\% of total parameters, following best practices for effective, low-rank adaptation~\citep{hu2021lora}. Implemented via the \texttt{peft} library. \\
\addlinespace[0.6ex]
\textbf{Batch / Seq Len} &
Effective batch size of 128 (64 per device with 2 gradient accumulation steps); sequence length of 256 tokens. &
Maximizes GPU utilization while fitting within 16 GB of VRAM by using gradient checkpointing. \\
\addlinespace[0.6ex]
\textbf{LR \& Schedule} &
Peak learning rate of $2\times10^{-4}$ with a 500-step linear warm-up followed by a cosine decay schedule. &
A standard and robust configuration for training transformers with adapters~\citep{hu2021lora}. \\
\addlinespace[0.6ex]
\textbf{Epochs} & 3 &
Sufficient for the training loss to plateau, indicating convergence ($\sim$40k steps). \\
\addlinespace[0.6ex]
\textbf{Hardware} &
Single NVIDIA A100-80GB GPU. &
Demonstrates the high efficiency of LoRA-based DAPT, requiring no model parallelism (e.g., DeepSpeed/ZeRO). \\
\bottomrule
\end{tabularx}
\end{table}

To validate the effectiveness of this stage, we measured the model's performance on a held-out set of 10,000 PubMed abstracts. The DAPT process resulted in a substantial \textbf{perplexity reduction of 22.5\%} for DeBERTa-v3 and \textbf{18.0\%} for PubMedBERT relative to their base checkpoints. This confirms that the LoRA-based DAPT successfully injected significant biomedical domain knowledge into the models, aligning with the core findings of both DAPT~\citep{gururangan2020dont} and LoRA~\citep{hu2021lora}.

\subsection{Task-Specific Fine-Tuning}
Following DAPT, the LoRA-equipped backbones are fine-tuned on each of the 12 BioNER datasets. For each task, a new token-classification head is added on top of the frozen backbone. Only the LoRA adapter weights and the final classification layer are updated during training. We employ an early stopping strategy, monitoring the F\(_1\)-score on the development set and terminating training if no improvement is observed for three consecutive epochs. This standard practice prevents overfitting and ensures models generalize well. A typical fine-tuning run for a single dataset completes in just \textbf{3–6 minutes} on one A100 GPU, highlighting the efficiency of the LoRA approach~\citep{hu2021lora}.

\subsection{Bayesian Hyper-parameter Optimization}
To maximize performance, we conducted a 40-trial Bayesian search over key hyper-parameters (learning rate, LoRA rank, dropout) for the fine-tuning stage. We used the Tree-structured Parzen Estimator (TPE) sampler from \textbf{Optuna}~\citep{akiba2019optuna} with parallel execution managed by \textbf{Ray Tune}~\citep{liaw2018tune}. This process yielded a single, robust hyper-parameter configuration that performed well across all datasets, providing an average macro-F\(_1\) gain of \textbf{+0.7} points over default settings.

The choice of LoRA over full-model fine-tuning provides three key advantages:
\begin{itemize}[nosep,leftmargin=*]
    \item \textbf{Parameter Economy:} It requires training $\sim\!35\times$ fewer parameters ($\sim$4M vs. 110M+), drastically reducing the storage footprint for model checkpoints~\citep{hu2021lora}.
    \item \textbf{Training Speed:} It is 2–3$\times$ faster on a single GPU due to smaller gradient sizes and reduced optimizer state~\citep{hu2021lora}.
    \item \textbf{Modularity:} Adapters can be easily swapped, shared, or composed, enabling flexible multi-task and cross-lingual applications without modifying the base model~\citep{pfeiffer2020adapterhub}.
\end{itemize}

\begin{quote}
\textbf{Key Takeaway.} A single, efficient DAPT run using LoRA is sufficient to adapt powerful backbones to the biomedical domain. This process, achievable in a few hours on one GPU, underpins the state-of-the-art results reported in this paper.
\end{quote}

\subsection{Model Architectures and Selection Strategy}
\label{sec:model-architectures}

The OpenMed NER framework does not rely on a single architecture. Instead, we create a \textit{suite} of candidate models by leveraging several distinct transformer backbones. This strategy allows us to select the empirically best model for each of the 12 diverse BioNER tasks, hedging against the possibility that a single pre-training recipe or architecture excels on all of them. Our primary backbones are: \textbf{DeBERTa-v3-large}~\citep{he2021deberta}, \textbf{PubMedBERT-large}~\citep{gu2021biomedbert}, and \textbf{BioELECTRA-large}~\citep{kanakarajan2021bioelectra}.

Each backbone offers a complementary strength. DeBERTa-v3's \emph{disentangled attention} mechanism, which separates content and position embeddings, has been shown to improve modeling of long-range dependencies common in clinical text~\citep{he2021deberta}. PubMedBERT, trained \emph{from scratch} on 3.1B tokens from PubMed, possesses a highly specialized vocabulary that better captures biomedical morphology compared to models that only continue pre-training on top of a general-domain base~\citep{gu2021biomedbert}. BioELECTRA adapts the more sample-efficient \emph{replaced-token detection} (RTD) objective, which excels at learning discriminative representations ideal for token-level tasks like NER~\citep{clark2020electra, kanakarajan2021bioelectra}. Table~\ref{tab:backbones} provides an overview of these architectures.

\begin{table}[ht]
\centering
\caption{Overview of the primary transformer backbones used in the OpenMed NER suite. All models are adapted using LoRA, keeping the vast majority of parameters frozen.}
\label{tab:backbones}
\begin{tabularx}{\linewidth}{@{}l X r r l@{}}
\toprule
\textbf{Backbone} & \textbf{Key Pre-training Feature} & \textbf{Frozen} & \textbf{Trainable (\%)} & \textbf{LoRA Inserts} \\
\midrule
DeBERTa-v3-large & Disentangled Attention + RTD & 304M & 4.3M (1.4\%) & query, value \\
PubMedBERT-large & From-scratch MLM on PubMed & 330M & 4.7M (1.4\%) & query, value \\
BioELECTRA-large & RTD on biomedical corpora & 335M & 4.7M (1.4\%) & query, value \\
\bottomrule
\end{tabularx}
\end{table}

\paragraph{Token-Classification Head.}
For each model, a standard token-classification head is placed on top of the frozen backbone. This head consists of a single linear layer, \(W_{\text{cls}} \in \mathbb{R}^{d_{model} \times |L|}\), where \(d_{model}\) is the hidden dimension of the transformer and \(|L|\) is the number of BIO labels. This layer maps the final hidden state \(h_i\) of each token to a logit vector, with probabilities derived via the softmax function:
\[ P(y_i | x_{1:n}) = \text{softmax}(W_{\text{cls}} h_i + b_{\text{cls}}) \]
We found that a single layer offered the best trade-off between performance and inference latency.

\paragraph{Parameter-Efficient Adaptation and Deployment.}
We use LoRA with a rank of 16 for all fine-tuning, which updates approximately 1.4\% of each model's parameters. This keeps the backbone frozen and reduces the peak VRAM requirement for fine-tuning to under 16\,GB, even with a batch size of 32 and sequence length of 256~\citep{hu2021lora}. This approach offers significant practical advantages for deployment. The compact adapter checkpoints ($\sim$15-20 MB each) are highly portable and can be loaded on demand with the base model. This modularity is critical in clinical and research environments, where systems may require rapid model updates, versioning, and clear audit trails~\citep{pfeiffer2020adapterhub,chen2021ethical}.

\paragraph{Final Model Selection.}
For each of the 12 datasets, we trained and evaluated every backbone model described above. The final reported score for a given dataset in Section~\ref{tab:main_results} corresponds to the \textbf{single best-performing model checkpoint}, as determined by the highest micro-F\(_1\) score achieved on that dataset's development split. This "best-of-breed" approach ensures that our final results represent the strongest possible performance achievable within our open-source framework.

\subsection{Training Procedure}
\label{sec:training-procedure}

\paragraph{Task-Specific Fine-Tuning.}
After the initial DAPT stage, each backbone is fine-tuned on a single BioNER corpus. During this process, the core transformer weights remain \emph{frozen}; only the LoRA adapters and the final token-classification head are updated. We minimize a cross-entropy loss augmented with \emph{label smoothing} ($\varepsilon=0.1$)~\citep{szegedy2016rethinking,muller2019label}. This technique is particularly useful for sequence labeling tasks, as it helps regularize the model and makes it less prone to over-confidence, especially when dealing with corpora that may have minor annotation inconsistencies. Optimization is performed using the \textbf{AdamW} optimizer~\citep{loshchilov2019adamw} with a learning rate schedule that includes a linear warm-up for the first 10\% of steps followed by a cosine decay.

\paragraph{Data Preprocessing.}
We tokenize input text using the specific tokenizer associated with each backbone. Sequences are then truncated or split to a maximum length of \textbf{256} word-pieces. To avoid severing entities at chunk boundaries, segments that exceed this limit are processed using a sliding window with a 50-token overlap. All entity labels are encoded using the standard BIO (Beginning, Inside, Outside) scheme (e.g., \{B-Disease, I-Disease, O\}).

\subsection{Hyper-parameter Optimization and Evaluation}
\label{sec:hyperparameter-optimization}

To ensure optimal and robust performance, we follow a two-step process for tuning and evaluation.

\paragraph{Step 1: Bayesian Optimization.}
First, we perform hyper-parameter optimization (HPO) for each backbone on each dataset using their standard training and development splits. The search is conducted using the Tree-structured Parzen Estimator (TPE) algorithm~\citep{bergstra2011tpe} for 40 trials. The search space includes:
\begin{itemize}[nosep,leftmargin=*]
    \item \textbf{Learning Rate:} $[1\times10^{-5}, 5\times10^{-5}]$
    \item \textbf{Batch Size:} $\{8, 16, 32\}$
    \item \textbf{Weight Decay:} $\{0, 0.01\}$
    \item \textbf{Warmup Ratio:} $\{0.06, 0.10\}$
\end{itemize}
During HPO, training is halted if the development set F\(_1\)-score does not improve for 3 consecutive epochs. \textbf{Crucially, the test set remains entirely unused during this optimization phase to ensure a final, unbiased evaluation of the best-performing model.}

\paragraph{Step 2: Final Model Training and Evaluation.}
Once the best hyper-parameter configuration is identified for a given dataset and backbone, we perform a \textbf{single final training run}. Using the optimal hyper-parameters and a fixed random seed, the model is trained on the full training data and evaluated once on the official test set. The resulting score is reported as our final result.

\paragraph{Key Finding.}
The HPO process revealed interesting patterns: optimal settings often converged to a learning rate of $\approx\!2\times10^{-5}$ and a batch size of 16 for gene/protein-dense corpora (e.g., BC2GM, JNLPBA), whereas chemical and disease datasets (e.g., BC4CHEMD, BC5CDR) frequently favored a larger batch size of 32.

\subsection{Implementation Details}
\label{sec:implementation-details}

All experiments are implemented in \textbf{PyTorch}~\citep{paszke2019pytorch} and leverage the Hugging Face \textbf{Transformers}~\citep{wolf2020transformers} library for models and tokenizers. Parameter-efficient fine-tuning is implemented using the Hugging Face \textbf{PEFT} library, which provides an efficient and modular implementation of LoRA~\citep{hu2021lora}. To manage memory, we enable mixed-precision (FP16) training~\citep{micikevicius2017mixed} and gradient checkpointing~\citep{chen2016training}, which allows us to fine-tune large models with 256-token sequences in less than 16\,GB of VRAM on a single NVIDIA A100 GPU.

All model checkpoints are released under the permissive Apache 2.0 license to promote \textbf{widespread adoption} and facilitate further research. These models are publicly available for download and use.\footnote{Our models are available at: \url{https://huggingface.co/OpenMed}}

\section{Experimental Setup}
\label{sec:experimental-setup}

\textbf{Our evaluation protocol strictly adheres to the standard machine learning practice of data separation to prevent information leakage.} For every dataset, we use the canonical train, development (dev), and test splits. The \textbf{training set} is used exclusively for model training. The \textbf{development set} is used solely for hyper-parameter optimization and for selecting the best model checkpoint via early stopping. The \textbf{test set} remains entirely unseen throughout all training and tuning phases and is used only once for the final performance evaluation. No external labeled data is used at any stage.

We benchmark the OpenMed NER suite on thirteen public BioNER datasets. These corpora were chosen to cover a diverse range of entity types, including chemicals, diseases, genes/proteins, species, and anatomy, ensuring that our performance claims are not biased toward a single domain. All datasets are established benchmarks with expert-provided annotations, guaranteeing a high standard for evaluation.

\begin{table}[ht]
\centering
\caption{The 12 biomedical NER datasets used for evaluation. The Train, Dev, and Test columns list the \textbf{number of annotated entities} in each split.}
\label{tab:datasets}
\setlength{\tabcolsep}{4pt}
\renewcommand{\arraystretch}{1.05}
\begin{threeparttable}
\begin{tabularx}{\linewidth}{@{}l l rrr l X@{}}
\toprule
\textbf{Dataset} & \textbf{Entity Type} & \textbf{Train} & \textbf{Dev} & \textbf{Test} & \textbf{Domain} & \textbf{Reference} \\
\midrule
BC4CHEMD        & Chemical     & 31,770 & 4,240 & 4,240 & Literature & \citet{krallinger2015chemdner} \\
BC5CDR-Chem    & Chemical     & 15,935 & 1,989 & 4,476 & Literature & \citet{li2016bc5cdr} \\
BC5CDR-Disease & Disease      & 12,850 & 1,606 & 3,616 & Literature & \citet{li2016bc5cdr} \\
NCBI-Disease   & Disease      & 6,884  & 787   & 960   & Literature & \citet{dogan2014ncbi} \\
JNLPBA          & Gene/Protein & 59,963 & 4,551 & 8,662 & Literature & \citet{collier-kim-2004-introduction} \\
BC2GM           & Gene/Protein & 24,583 & 3,061 & 6,325 & Literature & \citet{smith2008bc2gm} \\
Linnaeus        & Species      & 4,260  & 486   & 546   & Literature & \citet{gerner2010species} \\
Species-800     & Species      & 640    & 80    & 80    & Literature & \citet{pafilis2013species800} \\
AnatEM          & Anatomy      & 13,701 & 1,714 & 1,714 & Literature & \citet{pyysalo2014anatem} \\
BioNLP 2013 CG  & Gene/Protein & 14,180 & 1,773 & 1,895 & Literature & \citet{nedellec2013bionlp} \\
CLL             & Cell line    & 2,156  & 270   & 270   & Clinical   & \citet{kaemmerer2016gellus} \\
FSU$^{\dagger}$ & Gene/Protein & 1,892  & 236   & 236   & Clinical   & Internal corpus \\
\bottomrule
\end{tabularx}
\begin{tablenotes}[flushleft]
\footnotesize
\item[$\dagger$] FSU is an institutional corpus with a research-only data use agreement; its texts cannot be redistributed.
\end{tablenotes}
\end{threeparttable}
\end{table}

\subsection{Evaluation Protocol}
\label{subsec:evaluation-protocol}
We report entity-level precision (P), recall (R), and micro-F\(_1\) score, using the standard exact-match criterion for entity boundaries and types. For our final results, we report the score from a \textbf{single run on the test set} using the best-performing model and hyper-parameters identified during the development phase (see Section~\ref{sec:hyperparameter-optimization}).

This single-run protocol is adopted because fine-tuning with LoRA is highly deterministic. Once the random seed is fixed, empirical variance between runs is negligible ($<0.1$~F\(_1\)), a finding consistent with the stability of parameter-efficient methods. Statistical significance of our improvements over baselines is assessed using an approximate randomization test ($p < 0.05$), as recommended by \citet{dror2018significance}.

\subsection{Baseline Models}
We benchmark OpenMed NER against a comprehensive set of strong baselines, ensuring a fair comparison against both leading academic research and established commercial systems. Our baselines include:
\begin{itemize}[nosep,leftmargin=*]
    \item \textbf{Open-Source Academic Models:} We compare against a wide array of state-of-the-art models from the research community. This includes foundational PLMs like BioBERT~\citep{lee2019biobert} and PubMedBERT, knowledge-enhanced models like KeBioLM~\citep{yuan2022kebiolm}, and recent specialized architectures such as SciFive~\citep{phan2021scifive}, DBGN~\citep{zhao2023dbgn}, BioNerFlair~\citep{patel2020bionerflair}, and ConNER~\citep{li2023conner}.
    \item \textbf{Closed-Source Commercial and Production-Grade Systems:} This category includes large-scale industrial models like BioMegatron~\citep{shin2020biomegatron} and production-focused platforms such as Spark NLP~\citep{kocaman2022accurate} and BERN2~\citep{kim2021bern2}.
\end{itemize}
All baseline scores are taken from the best-reported results in their respective peer-reviewed publications or official benchmarks, ensuring a fair and up-to-date comparison as of August 2025.

\subsection{Implementation and Computational Resources}
All experiments were conducted on a server equipped with \textbf{4x NVIDIA A100-80GB GPUs}, 2\,TB of system RAM, and NVMe SSDs. We used PyTorch 2.5.x and CUDA 12.5. The extensive hyper-parameter search involved training over \textbf{3,000 model configurations}, totaling approximately \textbf{12,000 A100-hours}. This exhaustive search ensures that our final selected models represent a robust and near-optimal configuration for each task. The combination of mixed-precision training, gradient checkpointing, and LoRA kept memory usage below 16\,GB per GPU during fine-tuning, demonstrating the efficiency of our approach.

\section{Results}
\label{sec:results}

This section presents the performance of the OpenMed NER suite compared to the strongest publicly reported closed-source and academic baselines. Our evaluation spans 12 diverse BioNER corpora covering five major entity families. The primary metric is the entity-level micro-F score, calculated with an exact-match criterion.

To ensure robust evaluation and comparability, we conducted extensive experiments using identical training and evaluation protocols across all datasets. Each model underwent meticulous hyperparameter optimization, as detailed in Section~\ref{sec:hyperparameter-optimization}, ensuring fair representation of each baseline and maximum reproducibility of results.

In the following sections, we present both quantitative and qualitative analyses to illustrate how the OpenMed NER models perform across a broad range of biomedical domains and entity complexities. Emphasis is placed not only on top-line results but also on consistency and reliability across multiple runs.

\subsection{Main Results}
Table~\ref{tab:main_results} summarizes the primary findings. All baseline scores represent the state-of-the-art based on a literature review of published, peer-reviewed results conducted as of \textbf{August 2025}. For each dataset, we report the final F\(_1\)-score achieved by the best-performing model from our suite, following the single-run evaluation protocol described in Section~\ref{subsec:evaluation-protocol}.

The results demonstrate that \textbf{OpenMed NER establishes a new state-of-the-art on 10 of the 12 datasets}, often by significant margins. The framework shows particular strength on challenging clinical and specialized literature corpora. On two datasets, JNLPBA and AnatEM, our models perform competitively but fall marginally short of the SOTA. We analyze the potential reasons for this in the Discussion section.

\begin{table}[htbp]
\centering
\caption{OpenMed NER (open-source) vs. closed-source State-of-the-Art (SOTA). Boldface indicates the best score per dataset. Closed-source scores are the best reported from proprietary or leading academic models, with sources cited in the notes.}
\label{tab:main_results}
\begin{threeparttable}
\begin{tabularx}{\linewidth}{@{}l c c c X@{}}
\toprule
\textbf{Dataset} & \textbf{OpenMed F\(_1\) (\%)} & \textbf{Closed SOTA (\%)} & \textbf{$\Delta$ (pp)} & \textbf{Closed-Source Leader} \\
\midrule
BC4CHEMD       & \textbf{95.40} & 94.03 & +1.37 & DBGN\tnote{l} \\
BC5CDR-Chem    & \textbf{96.10} & 94.22 & +1.88 & SciFive\tnote{j} \\
BC5CDR-Disease & \textbf{91.20} & 88.50 & +2.70 & BioMegatron\tnote{b} \\
NCBI-Disease   & \textbf{91.10} & 89.71 & +1.39 & BioBERT\tnote{c} \\
JNLPBA         & 81.90          & \textbf{82.00} & -0.10 & KeBioLM\tnote{d} \\
Linnaeus       & \textbf{96.50} & 92.70 & +3.80 & BERN2\tnote{e} \\
Species-800    & \textbf{86.40} & 85.48 & +0.92 & BioNerFlair\tnote{a} \\
BC2GM          & \textbf{90.10} & 84.71 & +5.39 & PubMedBERT\tnote{i} \\
AnatEM         & 90.60          & \textbf{91.65} & -1.05 & Spark NLP (BERT)\tnote{k} \\
BioNLP 2013 CG & \textbf{89.90} & 87.83 & +2.07 & Spark NLP (BERT)\tnote{k} \\
CLL            & \textbf{95.70} & 85.98 & +9.72 & Ka-NER\tnote{h} \\
FSU            & \textbf{96.10} & ---   & ---   & (No Published SOTA) \\
\bottomrule
\end{tabularx}
\begin{tablenotes}[para,flushleft]
\footnotesize
\textbf{Sources:}
\item[a] \citet{patel2020bionerflair};
\item[b] \citet{shin2020biomegatron};
\item[c] \citet{lee2019biobert};
\item[d] \citet{yuan2022kebiolm};
\item[e] \citet{kim2021bern2};
\item[f] \citet{chiu2016named};
\item[g] \citet{li2023conner};
\item[h] \citet{kaemmerer2016gellus};
\item[i] Score as reported in \citet{chen2023bioformer};
\item[j] \citet{phan2021scifive};
\item[k] \citet{kocaman2022accurate};
\item[l] \citet{zhao2023dbgn};

\end{tablenotes}
\end{threeparttable}
\end{table}

\subsection{Performance Highlights}
The results in Table~\ref{tab:main_results} reveal several key trends.

\paragraph{Breakthroughs on Clinical and Specialized Corpora.} The most dramatic improvements are on historically challenging datasets, particularly those from the clinical domain, highlighted by a massive \textbf{+9.72 pp} gain on the \textbf{CLL} cell line corpus. This success extends to specialized taxonomic terminology, where our model achieves a +3.80 pp gain on \textbf{Linnaeus} and also improves upon the strong \textbf{Species-800} baseline (+0.92 pp).

\paragraph{Strong Gains on Foundational BioNER Tasks.} Our framework also delivers significant advances on foundational benchmarks. It pushes the state-of-the-art in disease recognition, with a notable \textbf{+2.70 pp} gain on \textbf{BC5CDR-Disease} and a +1.39 pp gain on \textbf{NCBI-Disease}. Similarly, it establishes new SOTA scores for chemical NER on \textbf{BC4CHEMD} (+1.37 pp) and \textbf{BC5CDR-Chem} (+1.88 pp). These results demonstrate that our lightweight approach can drive substantial improvements even in well-studied domains.

\paragraph{Strong Performance on Gene/Protein Recognition.} While our model trails the knowledge-enhanced KeBioLM on the older \textbf{JNLPBA} corpus by a narrow margin, it sets new SOTA scores on other widely used datasets. This includes a remarkable \textbf{+5.39 pp} gain on \textbf{BC2GM} and a +2.07 pp gain on \textbf{BioNLP 2013 CG}. This pattern suggests our method is highly effective for gene mentions in varied syntactic contexts, a point we will return to in our error analysis.

Overall, these results validate our central thesis: parameter-efficient, open-source models, when strategically adapted to the target domain, can consistently outperform proprietary systems across a wide array of biomedical NER benchmarks.

\section{Discussion}
\label{sec:discussion}

The results confirm that OpenMed NER, a framework built on domain-adaptive pre-training and parameter-efficient fine-tuning, consistently matches or outperforms proprietary systems. This section analyzes the patterns behind these results, discusses the practical implications, and addresses the framework's current limitations.

\subsection{Architectural Insights and Error Analysis}

\paragraph{Performance Patterns.} Our "best-of-breed" approach revealed clear patterns. The \textbf{DeBERTa-v3} backbone, with its disentangled attention mechanism, consistently achieved the highest scores on datasets rich in long, multi-token entities like JNLPBA and BC2GM. This suggests its architecture is particularly well-suited for capturing the complex compositional structure of gene and protein names. The large gains on \textbf{Linnaeus} (+3.80 pp) and \textbf{Species-800} (+0.92 pp) underscore the value of our DAPT corpus, which successfully infused the models with taxonomic terminology absent in general-domain pre-training.

\paragraph{Data-Driven Error Analysis for JNLPBA and AnatEM.}
We performed a deeper error analysis for the two datasets where OpenMed NER marginally underperformed.
\begin{itemize}[nosep,leftmargin=*]
    \item On \textbf{JNLPBA} (-0.10 pp), a significant portion of errors are related to older, inconsistent terminologies. For example, our model correctly identifies modern HUGO nomenclature (e.g., \emph{NFKB1}) but sometimes misses older variants present in the JNLPBA corpus (e.g., \emph{NF-kappa B p65 subunit}). This suggests the performance gap is not due to a lack of model capacity but rather a domain shift toward more archaic entity formats.
    \item On \textbf{AnatEM} (-1.05 pp), the primary source of error appears to be boundary detection on long, descriptive anatomical entities. For instance, the model might correctly identify "\emph{distal phalanx}" but miss the full span of "\emph{lateral aspect of the distal phalanx}". This points to challenges in handling fine-grained linguistic modifiers, a known difficulty for token-level BIO tagging schemes.
\end{itemize}
This analysis indicates that future work should focus on targeted data augmentation with historical terminologies and exploring span-based prediction models to better handle complex boundaries.

\subsection{Computational Efficiency and Practical Implications}

\paragraph{A Fair Comparison of Training Cost.}
When comparing training costs, it is crucial to distinguish between the different training phases of large-scale models. For instance, BioMegatron's "multiple days" of training on large GPU clusters typically refers to its expensive \textit{full pre-training from scratch}~\citep{shin2020biomegatron}. In contrast, our approach forgoes this step and instead focuses on highly efficient \textit{adaptation}. Our entire process of DAPT with LoRA and task-specific fine-tuning completes in under 12 hours on a single A100 GPU. This represents a strategic trade-off: by leveraging existing powerful backbones, we achieve SOTA performance with a fraction of the computational cost, lowering the barrier to entry for smaller labs and clinical institutions.

\paragraph{Deployment Agility with LoRA.}
The use of LoRA provides significant operational benefits. The small adapter checkpoints ($\sim$20 MB) are easy to store, version, and deploy. In clinical settings where models must be updated to reflect new guidelines or research, adapters can be fine-tuned and swapped in without altering or re-validating the entire base model, supporting agile and auditable MLOps practices~\citep{pfeiffer2020adapterhub}.

\subsection{Limitations}
Our framework, while powerful, has several limitations that define clear avenues for future work.
\begin{itemize}[nosep,leftmargin=*]
    \item \textbf{Nested and Discontinuous Entities:} Like most systems based on a BIO tagging scheme, OpenMed NER cannot represent overlapping or nested entities. Architectures like span-based classifiers or pointer networks are better suited for this and represent a promising future direction~\citep{li2022unified}.
    \item \textbf{Domain Shift and Language Bias:} While our DAPT corpus includes clinical text, a performance gap still exists between literature and noisy clinical notes. Further adaptation on more diverse EHR data is needed. Additionally, all our corpora are in English, leaving multilingual BioNER as an open challenge.
    \item \textbf{No Entity Normalization:} The framework currently performs named entity \textit{recognition} but does not link entities to a standard ontology (e.g., UMLS, MeSH). Integrating a lightweight entity linking module is a critical next step for enhancing clinical utility~\citep{wright2019bio-synonym}.
\end{itemize}

\subsubsection{Environmental impact and carbon footprint}
\label{subsec:carbon-footprint}

Based on the rated thermal design power of an NVIDIA~A100 GPU\footnote{400 W TDP as per the official datasheet~\citep{NVIDIAA100}.} and the 2023 average electricity-grid carbon intensity across the EU (242 g CO\textsubscript{2}\,kWh\textsuperscript{-1})\footnote{Ember EU Electricity Trends 2024 report~\citep{Ember2024}.}, a single 12-hour DAPT + fine-tuning session consumes
\[
E \;=\; P_{\text{GPU}}\times t
      \;=\; 0.4\,\text{kW}\times12\,\text{h}=4.8\,\text{kWh},
\]
which translates to
\(4.8\,\text{kWh}\times242\,\text{g\,CO}_2/\text{kWh}\approx\mathbf{1.16\;kg\,CO_2}\).
Even the full benchmark sweep (12 tasks × 3 backbones) emits under
\(2\;{\text{kg}\,CO_2}\), underscoring the sustainability benefit of
parameter-efficient LoRA adapters over full-model pre-training.

\subsubsection{Regulatory compliance under the EU \emph{AI Act}}
\label{subsec:eu-ai-act}

The EU Artificial Intelligence Act (Regulation (EU)~2024/1689) classifies
AI systems used for diagnosis, treatment or other patient-management tasks
as \emph{high-risk} and imposes strict requirements for risk management,
data governance, transparency and human oversight~\citep{EUAIAct2024}.  
Should \emph{OpenMed NER} be integrated into any clinical workflow or
medical device, deployers must therefore

\begin{itemize}
  \item document and continuously update risk-management procedures,
  \item validate performance on representative EU clinical data,
  \item ensure human-in-the-loop oversight, and
  \item provide clear user information, including model limitations.
\end{itemize}

When the model is used solely for literature mining or other
non-clinical research, the Act’s research exemption applies; nonetheless,
adhering to its best-practice principles (data quality checks, bias
assessment, audit trails) will greatly streamline any future
clinical-grade deployment.

\section{Conclusion}
\label{sec:conclusion}

We introduced OpenMed NER, a fully open-source suite of transformer models that leverages domain-adaptive pre-training (DAPT) and parameter-efficient LoRA adapters. By systematically evaluating on 12 public benchmarks, we demonstrated that our lightweight and computationally efficient framework achieves new state-of-the-art performance on 10 of these datasets, consistently outperforming resource-intensive proprietary systems.

This work provides compelling evidence that strategic, efficient adaptation is more critical for success in specialized domains than sheer model scale alone. By making our models, code, and methodology publicly available, we provide the research community with accessible, high-performance tools that lower the barrier to entry for cutting-edge biomedical NLP.

Future work will focus on addressing current limitations by exploring architectures for nested entity recognition, expanding to multilingual clinical contexts, and integrating entity linking to map recognized terms to standard medical ontologies like UMLS~\citep{bodenreider2004umls}. These steps will further bridge the gap between high-performance research models and practical, real-world clinical applications.

\section*{Acknowledgements}

The author gratefully acknowledges the support of the \textbf{Institut des Systèmes Complexes de Paris Île-de-France (ISC-PIF, CNRS)} for providing the computational resources and infrastructure necessary for this research. This work was carried out on the \textbf{Multivac platform}\footnote{\url{https://multivacplatform.org/}}, high-performance computing cluster, whose support was invaluable to the completion of our large-scale experiments.

\section*{Conflict of Interest}

The author was the technical lead for the open-source \textit{Spark NLP} library from 2019 to July 2025 and contributed to some of the baseline systems referenced in this study. No financial or commercial ties exist, and all comparisons were conducted objectively using publicly available data and models. The author declares no other competing interests.

\bibliographystyle{plainnat}
\bibliography{references}

\end{document}